\documentclass[final,3p,times,twocolumn,authoryear]{elsarticle}

\usepackage{amsmath, amssymb}

\journal{2024 European Conference of Computing in Construction}

\begin{document}

\begin{frontmatter}


\title{SCALABILITY IN BUILDING COMPONENT DATA ANNOTATION: ENHANCING FAÇADE MATERIAL CLASSIFICATION WITH SYNTHETIC DATA}
\author{
{Josie Harrison$^1$, Alexander Hollberg$^1$, and Yinan Yu$^1$}\\
{$^1$Chalmers University of Technology, Gothenburg, Sweden}\\
}

\title{}




\begin{abstract}
Computer vision models trained on Google Street View images can create material cadastres. However, current approaches need manually annotated datasets that are difficult to obtain and often have class imbalance. To address these challenges, this paper fine-tuned a Swin Transformer model on a synthetic dataset generated with OpenAI’s DALL-E and compared the performance to a similar manually annotated dataset. Although manual annotation remains the gold standard, the synthetic dataset performance demonstrates a reasonable alternative. The findings will ease annotation needed to develop material cadastres, offering architects insights into opportunities for material reuse, thus contributing to the reduction of demolition waste.

\end{abstract}




\end{frontmatter}


\label{}

\section*{Introduction} 
Demolition waste from construction and renovation activity is a growing problem internationally, making up 25\%-30\% of all waste generated in the EU \citep*{Anastasiou_GeorgiadisFilikas_Stefanidou_2014}. 
There is a general consensus that current practices have room for improvement for diverting and recovering materials from demolition waste \citep*{Kabirifar_Mojtahedi_Wang_Tam_2020}. 

Recovering demolition waste is not a common activity in many countries because there remain limitations that practitioners currently face. Not having insight into the 
availability of which specific materials will be available and at what time is often identified as a top barrier \citep*{Akanbi_Oyedele_Oyedele_Salami_2020}. The planning of 
projects that could use recovered materials typically begin several years before the actual construction begins. A material cadastre that maps detailed building material 
information for every building at the scale of a country with an estimate of when the materials could become available would significantly alleviate the current time window bottleneck.

Because of the significant potential of this solution, several researchers have turned their attention to developing city-scale material cadastres with computer vision 
models trained on images of building exteriors \citep*{Raghu_Bucher_DeWolf_2023, Arbabi_Lanau_Li_Meyers_Dai_Mayfield_DensleyTingley_2022}. A top pain-point in developing these models is acquiring the training data 
necessary for machine learning (ML). For deep learning models, some practitioners say that the size of training data should be 10x the number of weights in a network 
\citep*{Baum_Haussler_1988}—often leading to six figure digits and more.

Additionally, because the type of material is not known before the GPS coordinates are requested to collect training data, this process can lead to some materials (i.e. classes) having higher counts 
than other classes—also known as class imbalance. ML practitioners try to avoid class imbalance because a ML model will place importance on a class 
that has a higher chance of occurring in the training data. For example, if brick is the most common material in the training data, then the model is likely to predict brick 
when deployed in a real-world situation. This effect can be negligible if brick actually is the most common material in a city, but it is preferable to assume equal class 
counts since the true distribution of materials in a city is unknown.

The scale achieved by previous studies has been at the scale of a city; however, the blue sky vision for a tool like this would be at the scale of a country. 
This would mimic current practices of sourcing materials while staying within the bounds of maximum distance for a recovered material to travel before its sustainability 
benefits become negligible \citep*{Ginga_Ongpeng_Daly_2020}. There is some truth to the logic that if a model works at the scale of a city then it should have similar 
performance at the scale of a country; however, previous studies have identified that several problems can arise when scaling an image classification task 
\citep*{Maggiori_Tarabalka_Charpiat_Alliez_2017, Perronnin_Sánchez_Mensink_2010, Hendrycks_Basart_Mazeika_Zou_Kwon_Mostajabi_Steinhardt_Song_2022}. For example, wood siding 
can take on different colours and textures in different cities since it is often a regional material, but an image classification model may struggle to perform well if it hasn’t 
been trained on wood with a particular colour. Because it’s impossible to foresee all potential 
quirks that may arise with a given use case, it’s important to move beyond proof-of-concepts at the scale of a city and demonstrate that the model maintains reasonable 
performance at the scale of a country.

Furthermore, it would be desirable to test an image classification model's performance on residential and office interiors since this layer of a building typically has a high rate of change, resulting in a potentially high yield of reusable renovation waste. However, obtaining large and diverse datasets of building interiors can be challenging because of privacy concerns, data access limitations, and the labor-intensive process of data collection and annotation. As a result, public datasets for this task are limited in size, diversity, and representativeness, which may not fully capture the complexity and variability of real-world interior environments. Addressing the scarcity of this type of dataset could unlock an important layer in the quest for creating country-wide material cadastres.

Therefore, the aim of this paper is to investigate the potential of using synthetically generated images to augment and extend previous research done to classify façade 
materials in Google Street View (GSV) images. More specifically, the research questions that this paper addresses are: 
\begin{itemize} 
  \item What are the impacts on an image classifier’s performance when correcting the class imbalance of a manually annotated dataset with synthetic data? 
  \item Is the error distribution of an image classifier trained exclusively on synthetic data comparable to a classifier trained on manually annotated data when both are evaluated on a manually annotated test set? 
  \item Does a model trained and tested only with synthetic data have a similar error distribution to the manually annotated dataset?
\end{itemize}

The results from these research questions make several contributions to the field of urban mining for material cadastres. To the authors’ best knowledge, no previous studies have 
utilized synthetic images that mimic GSV images for augmenting and extending an image classification model to detect façade material. As a result, we present a novel approach 
to reduce the amount of time needed to develop a training dataset for the classification of façade materials, as well as provide an indication of the potential to use synthetic images to create datasets that would otherwise be difficult to obtain. Additionally, we present two methods to decrease errors in the current 
approaches: utilizing a higher resolution model by reducing the problem space and augmenting a manually annotated dataset to correct class imbalance. The workflow developed 
in this research can be extended to detect façade materials that have no manually annotated training data and the workflow can be used to improve performance of the current 
state-of-the-art for this specific use case.

\section*{Background}

\subsection*{Material stock datasets}
In 2023, researchers from ETH Zurich published the first urban-scale ground truth dataset—hereon referred to as the Urban Resource Cadastre (URC) dataset—to detect façade 
materials in Tokyo, New York City, and Zurich \citep*{Raghu_Bucher_DeWolf_2023}. The researchers collected this data by first identifying GPS coordinates of 
interest, requesting the corresponding GSV image for these GPS coordinates, and manually annotating all resulting images. This dataset contains 971 annotated 400x600 pixel 
images with the labels ‘brick’, ‘stucco’, ‘rustication’, ‘metal’, ‘siding’, ‘wood’, ‘null’ (for images with no façade), and ‘other’ (for façade material that did 
not belong to any of the other labels). However, the authors noted that collecting this type of dataset can easily become a time-consuming process.

\subsection*{Generating images for training data}
In recent years, there were significant developments with synthetic images generated for the purpose of training downstream models. Research in this area is often fueled by 
use cases where acquiring more data is time/cost prohibitive, it is necessary to mask personally identifiable information for privacy, or it is not possible to collect data 
at all \citep*{Man_Chahl_2022}. There exists a wide variety of generative text-to-image models; however, it has recently been accepted that diffusion models outperform 
previous text-to-image generative models. Inspired by non-equilibrium thermodynamics, diffusion models replicate the diffusion process observed in physics by teaching a deep 
learning model to add and reverse noise in images \citep*{Sohl-Dickstein_Weiss_Maheswaranathan_Ganguli_2015}. 

The lack of diversity within a synthetic dataset is a known limitation associated with generating images for training models. To overcome this, \cite{He_Sun_Yu_Xue_Zhang_Torr_Bai_Qi_2023} 
used “language enhancement” to add variety to the prompts given to the generative model while also using a filter to discard images that didn’t resemble the target image. 
The “language enhancement” involved using a simple keyword-to-sentence Natural Language Processing (NLP) model based on the T5 model \citep*{He_Sun_Yu_Xue_Zhang_Torr_Bai_Qi_2023} 
to make sentences from keywords. This method to diversify the generated output is prone to straying far from the desired image, however. To counteract this effect, 
\cite{He_Sun_Yu_Xue_Zhang_Torr_Bai_Qi_2023} proposed a “CLIP Filter” that leveraged CLIP zero-shot classification confidence. However, the authors aimed to create a fully 
automated process, which could be excessive for use cases where there is enough value in just generating synthetic images alone—which is this paper’s use case.
\subsection*{Façade materials with high reusability potential}
At this point, it becomes apparent that we should justify the selection of specific façade materials for the synthetic generation of images. For example, there exist buildings 
built with ETFE plastic bubbles, but since these projects are rare, their inclusion in a training dataset would confuse a ML model. While research in the area 
of the reusability potential of specific building materials is rather sparse, \cite{Icibaci_2019} and \cite{Iacovidou_Purnell_2016}  identified stone with lime-based mortar, curtain walls, and concrete panels as materials with high reusability potential—hereon referred to as the High Reusability Potential (HRP) labels. 

\subsection*{Image classification}
Originally developed for NLP tasks \citep*{Vaswani_Shazeer_2017}, vision transformers (ViTs) are deep learning models that take 
an image separated into patches (with its positional information saved), encode these patches into a memory, and compare that memory with a target value 
\citep*{Dosovitskiy_Beyer_Kolesnikov_2021}. The memory is created with ‘attention’, which maps all the patches to each other to establish relative importance between them 
all \citep*{Vaswani_Shazeer_2017}. The Swin Transformer is a variant of the ViT methodology, which has shown to outperform many other models on image classification tasks 
while also being conscious of computational efficiency \citep*{Liu_Lin_2021}.  

\begin{figure*}[h!]
	\centering
  	\includegraphics[width=\textwidth]{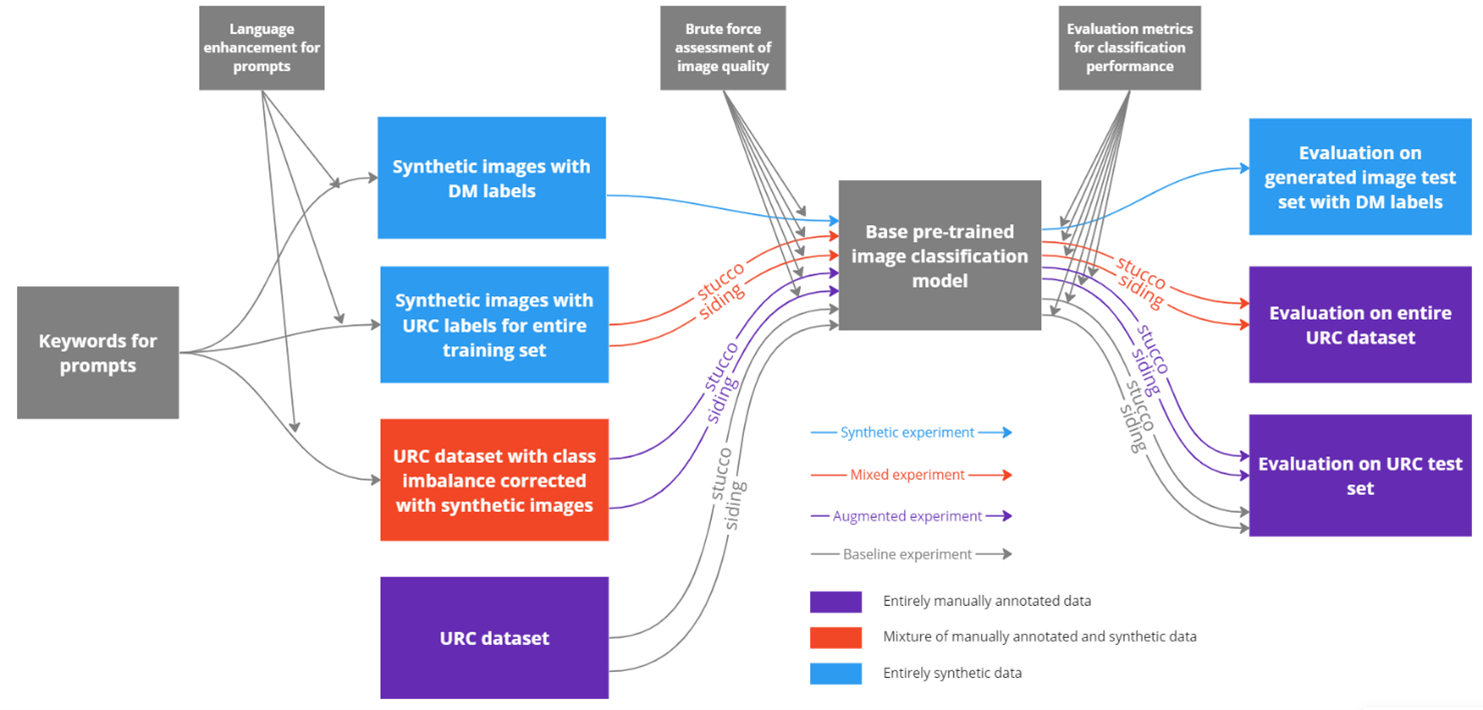}
  	\caption{Workflow showing all four experiments: baseline, augmented, mixed, synthetic with sub-experiments for stucco and siding.}
  	\label{fig:Workflow}
\end{figure*}
 
\section*{Experiment}
\subsection*{Workflow}
The workflow shown in Figure~\ref{fig:Workflow} assembled all materials and methods together into four experiments. There are three main experiments compared to a baseline 
experiment: augmented, mixed, and synthetic. For each experiment, a pre-trained image classification model was fine-tuned with the respective dataset, with synthetic images 
first passing through a brute-force image quality evaluation. After model training, the predictions were evaluated on the experiment’s test dataset. The synthetic experiment’s 
test set is entirely synthetic while the other three experiments use a manually annotated test set. 

\subsection*{Materials}
\begin{figure}[ht] 
	\centering
	\includegraphics[width=\linewidth]{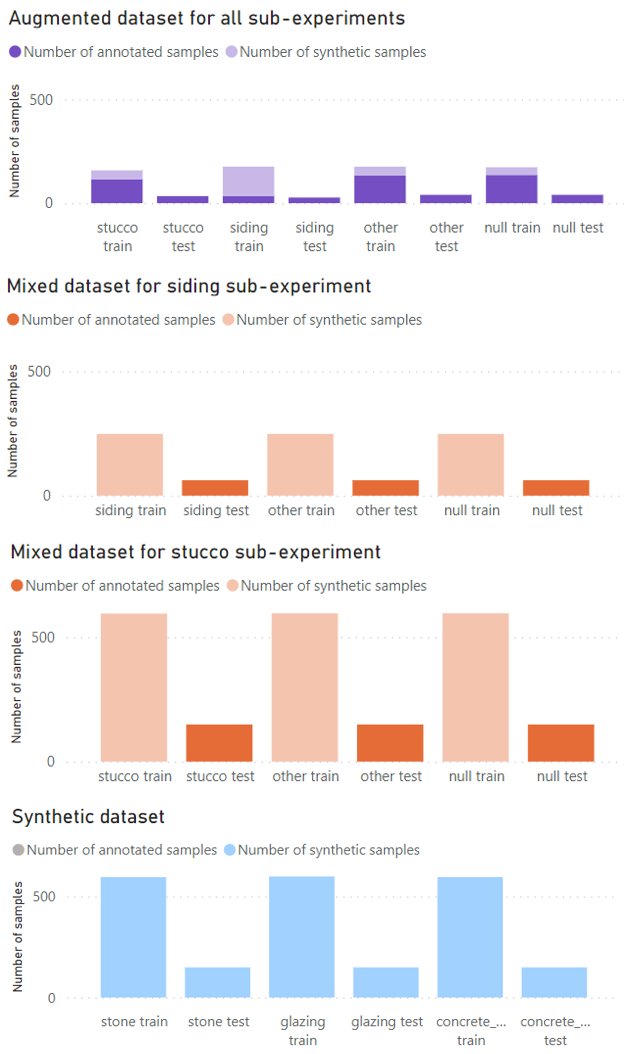}
	\caption{Class distributions for the different datasets.}
	\label{fig:DatasetDistribution}
\end{figure}

Drawing on previous research, four datasets (baseline, augmented, mixed, and synthetic) were collected for the purposes of training an image classification model to replicate 
previous work and to also train a model on unseen labels. The baseline, augmented, and mixed datasets used the URC labels, while the synthetic dataset used the HRP labels. For 
the augmented, mixed, and synthetic experiments (shown in Figure~\ref{fig:DatasetDistribution}), varying amounts of 512x512 pixel synthetic images were generated. This 
resolution was suitable for image classification models that can be trained on consumer-grade hardware.

For the baseline, augmented, and mixed experiments, the number of labels was reduced to three to create sub-experiments: null, other, and one label of interest in order to 
reduce the parameter space of the image classification model. ‘Stucco’ and ‘siding’ were isolated as two labels of interest because ‘stucco’ had a similar class distribution 
as the ‘null’ and ‘other’ labels, and because ‘siding’ had a high class imbalance in the original dataset. Therefore, there were two sub-experiments for the baseline, 
augmented, and mixed experiments using two sets of labels: “null, other, stucco” and “null, other, siding”. 

As mentioned earlier, the URC dataset came with class imbalance, so the augmented experiment corrected this class imbalance with synthetic images. The mixed experiment used 
only synthetic images for training while leaving the entire URC dataset as the test dataset. Finally, the synthetic experiment trained and tested entirely on synthetic images 
using the HRP labels: ‘stone’, ‘curtain wall’, and ‘concrete panels’. The synthetic experiment dataset size was made to match the mixed stucco experiment’s dataset set to 
maintain comparability. 

\vspace*{-6pt}
\nointerlineskip 

\subsection*{Methods}
There are several open source versions of diffusion models that are capable of creating synthetic images; however, the effort required to get them into an operational state can often be time-prohibitive while also 
requiring intensive hardware. For that reason, OpenAI’s DALL-E 2 model was chosen to generate images because it is the most effective to leverage with just a simple API request. A portmanteau of the famous surrealist artist Salvador Dalí and the Pixar character WALL-E, DALL-E takes in a user defined text prompt to create highly realistic images. The DALL-E 2 text-to-image generation model was trained on 650 million images \citep{DALL-E_2022}; however, OpenAI does not disclose the source of training data for any of their models \citep{AI_Art_Impact_2023}.
To reduce this dependency on an external organization, it would be ideal to develop an open source model in-house, however.

A limitation of generating synthetic GSV images was the inability for DALL-E to create believable images with more than one façade material. Other researchers found 
similar difficulties when requesting more than three components with descriptions about shape or colours \citep*{Marcus_Davis_Aaronson_2022}. Additionally, it was 
possible to utilize a higher resolution image classification model by reducing the label set to one material of interest combined with the ‘null’ and ‘other’ labels. For 
these reasons, it was decided to generate images with one label. Since this paper’s use case would not require predicting several labels in real-time, it is 
proposed that the implementation would require several models with each model focusing on one label (as well as the non-buildings and other materials).

\begin{figure}[ht] 
	\centering
	\includegraphics[width=\linewidth]{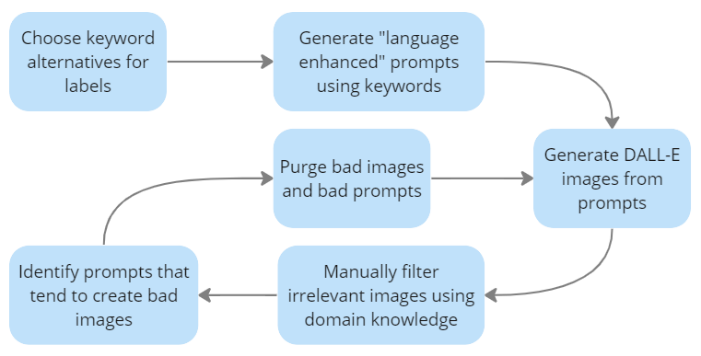}
	\caption{Brute force method to filter irrelevant synthetic images from accepted images.}
	\label{fig:BruteForceMethod}
\end{figure}

As mentioned earlier, there was enough value in creating the synthetic images alone to not utilize a fully automated filtering process. Therefore, a brute force version of 
this process was suitable. This brute force method shown in Figure~\ref{fig:BruteForceMethod}  involved generating a collection of prompts from keywords, generating a few 
hundred images by randomly sampling a prompt from the collection, manually removing irrelevant images from the final dataset, and taking note of the few prompts that were 
likely to produce the best images. The remaining dataset was generated using this final collection of ‘high batting average’ prompts.

To achieve a classification model that can perform at the scale of a country, it is desirable to encourage diversity in the synthetic images by injecting additional keywords into the prompt. For this experiment, the facade material label was given possible synonyms, a time period when the façade material was likely to be common, and cities where the facade material can be found. However, these keywords could be extended to any aspect of a building, such as type, size, or neighbourhood density. For example, ‘siding’ was assigned the 
synonyms 'shiplap', 'feather edge', 'fiber cement' and the time period '20th century'. For all labels in the augmented and mixed experiments, the original URC 
dataset cities (New York City, Zurich, and Tokyo) were used as keywords. For the synthetic experiments, the cities 'Vancouver', 'San Francisco', and 'Amsterdam' were chosen arbitrarily but would benefit from additional cities in future experiments.

For the image classification task, it was decided to use a Swin Transformer model pre-trained on the ImageNet-21k dataset to maximize comparability to previous studies and 
to leverage a well-known image classification model. A pre-trained image classification model is trained on a large generic dataset with the intention to later 'fine-tune' the 'head' of the model on tailored datasets. This method generally improves accuracy and reduces the need for a large tailored dataset. The ImageNet-21k dataset is commonly used for pre-training because it is 14 million images with 21,000 labels and contains a variety of natural images. \cite{Raghu_Bucher_DeWolf_2023} achieved a macro-averaged F1 score of 0.93 by applying the pre-trained Swin Transformer v2 model 
to the URC dataset. However, the authors used the model version with a resolution of 192 pixels, which was not adequate for the synthetic image dataset. It was seen that the 
synthetic images struggled with creating the same texture detail as real images, and this data loss was compounded when used with a low-resolution model. This is akin to 
taking a photocopy of a photocopy. Therefore, the experiments in this study use the Swin Transformer v2 model with 384 pixel resolution.

\vspace*{-6pt}
\nointerlineskip 

To provide a quantitative and objective measurement of the image classification performance, it was decided to calculate the weighted F1 score, precision, recall, as well as 
create a receiver operating characteristic (ROC) graph and confusion matrices for each experiment. The confusion matrix is a grid of all target labels with the y-axis 
representing the true test values and the x-axis representing the predicted test values. In this way, it’s possible to index how many samples were predicted for a specific 
label: the first cell in the grid shows the number of samples predicted as ‘null’ were correctly classified. It is not possible to decide which dataset performed the best by 
looking at these confusion matrices, however, so quantitative metrics were calculated: precision and recall.

\begin{equation}\label{eq:precision}
  precision = \frac{tp}{tp + fp}
\end{equation}

\begin{equation}\label{eq:recall}
  recall = \frac{tp}{tp + fn}
\end{equation}

Precision \eqref{eq:precision} refers to the likelihood that a prediction will be correct, while recall \eqref{eq:recall} refers to the likelihood that all samples will be 
correctly classified. In its basic binary form, the precision is the true positives (\textit{tp}) divided by all samples marked as positive (\textit{tp + fp}) \eqref{eq:recall}. 
Similarly, the recall formula \eqref{eq:recall} divides the true positives (\textit{tp}) by the sum of true positives and false negatives (\textit{tp + fn}). The F1 score 
\eqref{eq:f1score} provides a harmonic mean of both precision and recall.

\begin{equation}\label{eq:f1score}
  F = \frac{2 * precision * recall}{precision + recall}
\end{equation}

When working with multi-class problems, it is necessary to choose an averaging mechanism for the F1 score, precision, and recall metrics. The ‘weighted’ average is recommended 
for imbalanced classes because it assigns different weights to each class based on their prevalence in the dataset. This prevents the dominant class from strongly influencing 
these metrics. 

Lastly, the ROC curve graph was created to visually compare the performance of all four experiments. In this study’s context, the ROC curve can be used to visualize the 
characteristics of the performance; at what threshold between the true positive rate and false positive rate the model becomes uncertain about its predictions.

\section*{Discussion and result analysis}
\subsection*{Image generation}
As with many NLP models, the generated prompts contained a fair amount of hallucination; however, these nonsensical prompts surprisingly created some of the best images. For 
example, a successful prompt for the ‘stone’ label was, "The construction of building made of stone is in the medium of the Artistic Gymnastics." It was valuable to refrain 
from editorial oversight of the initial prompts to allow for spontaneity in the generated images. Because of this, the resulting datasets had more diversity than if  
one prompt created all the synthetic images.

Some labels had a higher number of generated images flagged as irrelevant than other labels, with an average of 24\% of all images being flagged as irrelevant. In most cases, 
the irrelevant images featured materials different from the target label. For example, the ‘stucco’ label often created images showing metal, siding, or rustication, which 
resulted in 45\% of all ‘stucco’ images being flagged as irrelevant. In some cases, the generative image model created images with no building shown at all. 

\begin{figure}[h!] 
	\centering
	\includegraphics[width=\linewidth]{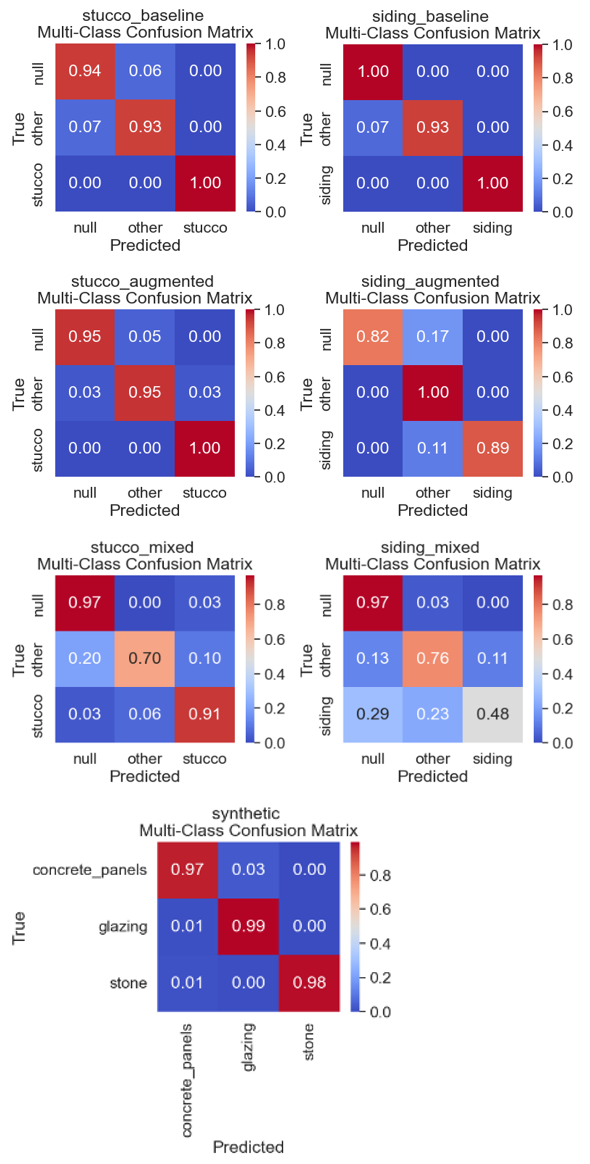}
	\caption{Confusion matrices for all experiments.}
	\label{fig:ConfusionMatrices}
\end{figure}

\subsection*{Image classification}
By reviewing the confusion matrices, ROC curve graph, and bar graph comparing F1 scores, precision, and recall, we are able to achieve a nuanced and holistic evaluation of the 
image classification results in the context of the research questions. 

The confusion matrix compares the number of true labels versus predicted labels. To aid in comparability and interpretability, the numbers of each row in the matrix were 
normalized to a percentage scale of 0\% to 100\%. The set of confusion matrices (Figure~\ref{fig:ConfusionMatrices}) show which labels perform the worst and best for each experiment and whether any degradation of performance 
is acceptable. Comparing the augmented stucco experiment to the baseline, it can be seen that the augmented dataset offered a slight advantage for the ‘null’ and ‘other’ 
labels, while maintaining perfect performance for the ‘stucco’ label. When using the mixed dataset, there was decrease in performance (100\% of true ‘stucco’ samples predicted 
as ‘stucco’ versus 91\% of true ‘stucco’ samples predicted as ‘stucco’). Moving to the siding experiment set, the augmented siding experiment had a decrease in performance 
(as opposed to the augmented stucco experiment). This is likely because the ‘siding’ augmented class had a higher proportion of synthetic to manually annotated images, which 
triggers the question that there might be a threshold where the ratio of synthetic to manually annotated images hinders performance. The mixed siding experiment showed further degradation with only 48\% of predicted ‘siding’ 
samples matching the true ‘siding’ label. This may have been caused by the size of the mixed siding dataset (248 training samples and 62 testing samples for each class), which 
was half the size of the mixed stucco dataset (596 training samples and 149 testing samples for each class). Lastly, it is no surprise that the synthetic experiment achieved 
near perfect results since the model had no way to test its performance on ‘in the wild’ images. Comparing the synthetic experiment results to the mixed experiment offers 
insight into the possible ways a model trained and tested on purely synthetic data might degrade when brought into a real data scenario.

The ROC curve graph is helpful in comparing each experiment’s discrimination capability between different classes. A model with a curve that hugs closely toward the top left 
corner suggests high performance in sensitivity and specificity, while the straight, black, dotted line represents the performance of a random classifier. A model with a ROC 
curve too close to this random classifier line indicates that the model is essentially predicting a class at random.

\vspace*{-6pt}
\begin{figure}[ht] 
	\centering
	\includegraphics[width=\linewidth]{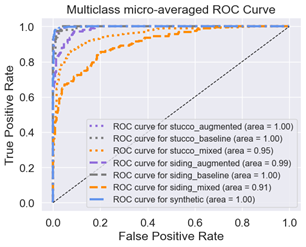}
	\caption{ROC curve graphs comparing the stucco and siding sub-experiments.}
	\label{fig:ROCcurves}
\end{figure}
\nointerlineskip 
\vspace{12pt}
From the ROC curves (Figure~\ref{fig:ROCcurves}), it can be seen that there are a few experiments that achieved both high sensitivity and specificity across all thresholds, 
indicated with an Area Under the Curve (AUC) value of 1.00: both baseline experiments, the augmented stucco experiment, and both synthetic experiments. The two mixed 
experiments achieved AUC values of 0.95 and 0.91 for stucco and siding, respectively, which is generally considered quite a good score. Since those ROC curves are positioned 
well above the random classifier line, it can be interpreted that these mixed models achieved a meaningful separation between true positives and false positives.  

Evaluating the weighted precision, recall, and F1 score of the four experiments provides a global view of the overall performance of each experiment. A higher precision 
indicates a lower rate of false positives, a higher recall indicates a lower rate of false negatives, while a higher F1 score indicates a balance between precision and recall.

\begin{figure}[ht] 
	\centering
	\includegraphics[width=\linewidth]{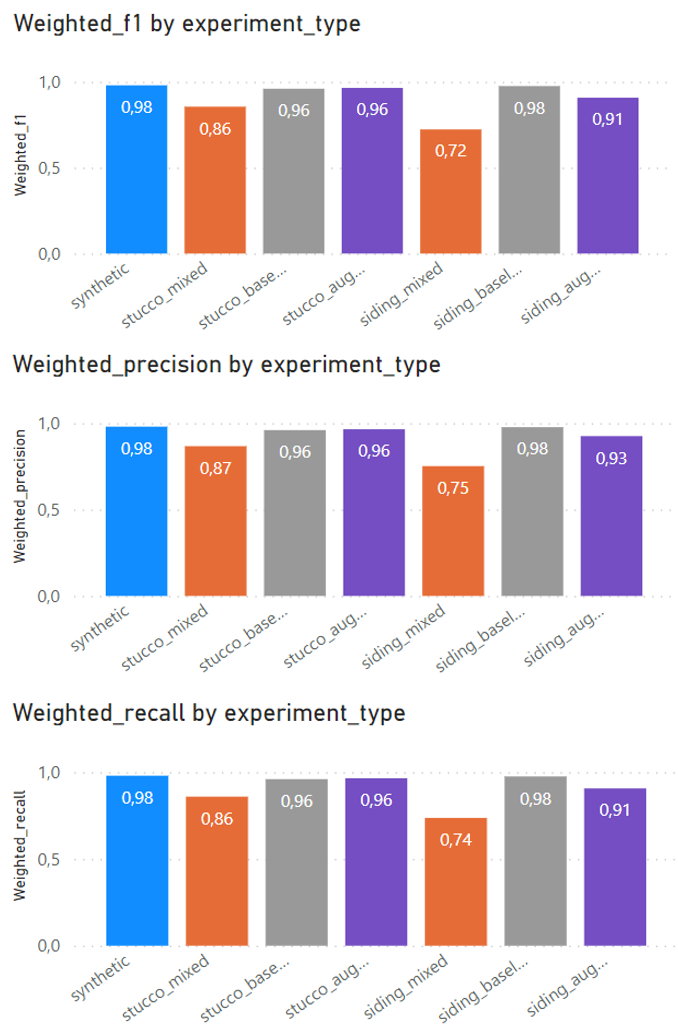}
	\caption{Weighted F1, precision, and recall for all experiments.}
	\label{fig:GlobalMetrics}
\end{figure}

The bar graphs showing the weighted F1 score, precision, and recall (Figure~\ref{fig:GlobalMetrics}) further supports the observations that models trained with some 
augmentation can increase performance and models trained with synthetic images maintain reasonable results on a manually annotated test dataset. The highest performing 
experiments were the synthetic experiment and the baseline siding experiment, achieving a value of 0.98 for all three metrics. Although the confusion matrices differed 
slightly, the baseline stucco and augmented stucco experiments achieved the same score of 0.96 for all three metrics. The augmented siding experiment performed slightly 
worse than the baseline siding experiment with a value of 0.91 for F1 score and recall (and a value of 0.93 for precision). This further supports the finding that too much 
augmentation can degrade performance. Although the mixed experiments had a decrease in metrics, the mixed stucco experiment maintained a reasonable performance with the 
values 0.86, 0.87, 0.86 for F1 score, precision, and recall. The mixed siding experiment showed further degradation with values 0.72, 0.75, 0.74 for F1 score, precision, 
and recall. However, the training and testing sets were half the size of the mixed stucco experiment, which could have influenced performance. 
 
\vspace*{-6pt}
\nointerlineskip 
\vspace{12pt}
There were several key findings from this study; some that offer an improvement over the state-of-the-art, and some that provide a nuanced analysis into the trade-offs made 
when training a model on synthetic images. In terms of overall improvement, this study achieved a higher F1 score, precision, and recall than the state-of-the-art by utilizing 
a higher resolution model (Swin Transformer v2 at 384 pixel resolution) and reducing the problem space to three labels (null, other, and a label of interest). This study 
achieved a 0.98 weighted F1 score, 0.98 recall, and 0.98 precision on the baseline siding experiment while the state-of-the-art achieved 0.93 macro F1 score, 0.91 recall, 
and 0.96 precision \citep*{Raghu_Bucher_DeWolf_2023}. For marginal improvements, it was found that augmenting a dataset with 27\% synthetic data (for the ‘stucco’ class) offered a slight 
improvement over the baseline. This had the opposite effect when using a dataset with 80\% synthetic data (for the siding class). As for the possibility to train a model on 
entirely synthetic data, the performance was generally worse than the baseline, but still a reasonable performance with the mixed stucco experiment capturing 91\% of true 
positives in the confusion matrix, achieving an AUC value of 0.95, and scoring 0.87 for precision and 0.86 for the F1 score and recall. The results from the mixed siding 
experiment may have been adversely affected by the size of the dataset (62 samples for test and 248 samples for training); potentially signaling a need to have a larger 
dataset when training on synthetic images. The near perfect results of the synthetic experiment show that it is possible to train a model on synthetic images for a new label 
set, but demonstrates the value of testing a model on manually annotated images.
 
\section*{Conclusions}
Recent advancements in text-to-image generation have made it feasible to train image classification models on synthetic images, addressing previous challenges of data scarcity and enabling scalability to a national or global level for façade material identification. In this study, we explored the potential benefits and trade-offs that could be made when using varying amounts of synthetic images for training a classification model, as well as proposed an improvement in 
overall performance with a higher resolution model. 
The research contributes a novel approach in urban mining for material cadastres by utilizing synthetic images to augment and extend an image classification model for detecting façade materials, offering a potential solution to reduce training dataset development time and address data scarcity challenges while improving current approaches' performance.
This, of course, comes with caveats: the text-to-image prompts should trigger diversity in the generated images, the model should be 
at least 384 pixel resolution, the problem space should be reduced to three labels (null, other, and label of interest), the synthetic training data should be large enough, 
and it is worthwhile to still manually annotate a test set to develop an intuition for the types of errors that will happen. The results from this study are limited to replicating GSV images for the study’s specific set of labels and would not necessarily guarantee reasonable performance for synthetic images of building interiors or niche facade materials (such as ceramic tile cladding). Access to a high-performance GPU (an nVidia GeForce RTX 3070 8GB GPU) enabled utilization of the pre-trained Swin Transformer v2 384 pixel resolution model, which may not be possible to train on a CPU alone. Additionally, there could be privacy concerns associated with remotely compiling building material inventories, particularly regarding historical (materials that should not be moved from their location) or high-security façade materials (materials that may reveal structural vulnerabilities).

These limitations indicate future research that could progress the field of urban mining for material cadastres further. While this experiment achieved good results mimicking GSV images, it would not be safe yet to conclude that any type of building image could be replicated with synthetic images. Therefore, it would be valuable for future work to explore applying the workflow to interior office and residential synthetic images. For the resolution of synthetic images, there was a considerable difference in performance when moving from the 192 pixel model to the 384 pixel 
model resolution, which begs the question whether generating higher resolution images and training with an even higher resolution model may increase performance further. When generating the synthetic images, a brute-force method was used to separate 
irrelevant images from the training set because it was suitable for this use case; however, an automatic filtering method would greatly reduce the time needed to prune the 
output of the generative model. Further studies using synthetic images could benefit material cadastre research by unlocking building interior datasets, by increasing performance of image classification models even more, and by reducing the time needed to filter synthetic images. Developing a robust image classification model that has proven reliable performance in detecting building materials in a wide variety of contexts, scales, and building layers would be an important milestone in the goal to develop country-wide material cadastres for reusable demolition and renovation waste.

\section*{Acknowledgments}
This project was financially supported by the Swedish Foundation for Strategic Research.


 
 \bibliographystyle{elsarticle-harv} 
\bibliography{references}





\end{document}